\documentclass[10pt,twocolumn,letterpaper]{article}
\usepackage{iccv}              % To produce the CAMERA-READY version
\usepackage{booktabs}
\usepackage{multirow}
\usepackage{algorithm}
\usepackage{algorithmic}

\definecolor{iccvblue}{rgb}{0.21,0.49,0.74}
\usepackage[pagebackref,breaklinks,colorlinks,allcolors=iccvblue]{hyperref}

\def\paperID{10808} % *** Enter the Paper ID here
\def\confName{ICCV}
\def\confYear{2025}

\def\red{\textcolor{red}}

\title{ViewSRD: 3D Visual Grounding via Structured Multi-View Decomposition}

\author{Ronggang~Huang$^{1*}$, Haoxin~Yang$^{1*\dagger}$, Yan~Cai$^1$, \\ Xuemiao~Xu$^{12345\dagger}$, Huaidong Zhang$^{1}$, Shengfeng~He$^6$. \\
\small{$^1$ South China University of Technology. }
\small{$^2$ Guangdong Engineering Center for Large Model and GenAI Technology.} \\
\small{$^3$ State Key Laboratory of Subtropical Building and Urban Science.}
\small{$^4$ Ministry of Education Key Laboratory of Big Data and Intelligent Robot.}  \\
\small{$^5$ Guangdong Provincial Key Lab of Computational Intelligence and Cyberspace Information.}
\small{$^6$ Singapore Management University.}
}

\begin{document}
\maketitle

\renewcommand{\thefootnote}{\fnsymbol{footnote}}
\footnotetext[0]{Accepted by ICCV 2025.}
\footnotetext[1]{The first two authors contributed equally.}
\footnotetext[2]{Corresponding authors:xuemx@scut.edu.cn, harxis@outlook.com.}

\begin{abstract}
3D visual grounding aims to identify and localize objects in a 3D space based on textual descriptions. However, existing methods struggle with disentangling targets from anchors in complex multi-anchor queries and resolving inconsistencies in spatial descriptions caused by perspective variations.
To tackle these challenges, we propose ViewSRD, a framework that formulates 3D visual grounding as a structured multi-view decomposition process. First, the Simple Relation Decoupling (SRD) module restructures complex multi-anchor queries into a set of targeted single-anchor statements, generating a structured set of perspective-aware descriptions that clarify positional relationships. These decomposed representations serve as the foundation for the Multi-view Textual-Scene Interaction (Multi-TSI) module, which integrates textual and scene features across multiple viewpoints using shared, Cross-modal Consistent View Tokens (CCVTs) to preserve spatial correlations. Finally, a Textual-Scene Reasoning module synthesizes multi-view predictions into a unified and robust 3D visual grounding.
Experiments on 3D visual grounding datasets show that ViewSRD significantly outperforms state-of-the-art methods, particularly in complex queries requiring precise spatial differentiation. Code is available at \url{https://github.com/visualjason/ViewSRD}.

\end{abstract}
    
\vspace{-15pt}
\section{Introduction}

\begin{figure}[t]
    \centering
    \includegraphics[width=1.0\linewidth]{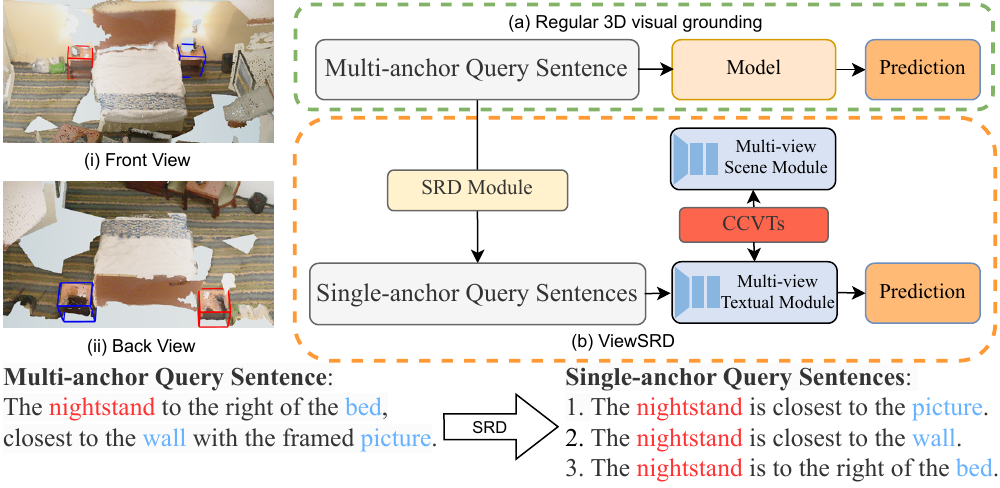}
    \caption{(a) Previous 3DVG methods struggle with ambiguities from complex multi-anchor queries and perspective shifts. (b) ViewSRD addresses this by using the SRD module to simplify queries and the CCVTs to capture viewpoint variations in both scene and textual modal, boosting cross-modal feature interaction and enhancing grounding accuracy.
    }
    \label{fig1}
    \vspace{-15pt}
\end{figure}

3D Visual Grounding (3DVG) aims to establish semantic correspondences between natural language descriptions and target objects in a 3D space~\cite{xu2024multi, huang2024structure}. This task has gained significant attention in applications such as visual language navigation~\cite{zhou2024navgpt,li2024panogen}, intelligent agents~\cite{yang2023appagent,calisto2023assertiveness}, and autonomous vehicles~\cite{cui2024drive,feng2023dense}.

Traditional single-view approaches rely on 2D sampled images to extract scene information~\cite{talebirad2023multi} or construct scene graphs from textual descriptions~\cite{wu2023eda}. However, these methods are inherently limited by their dependence on single-view cues, as language descriptions often presuppose specific viewpoints. To overcome this limitation, recent research has explored multi-view 3DVG, integrating multiple perspectives to enhance robustness~\cite{choudhary2024talk2bev,lyu2025mmscan,zhu2024unifying}. Some methods process distinct descriptions for different viewpoints via manual annotation and learning~\cite{shi2024aware,guo2023viewrefer}, while others incorporate spatial modules to encode relative spatial coordinates under specific perspectives~\cite{chang2024mikasa}. However, they typically address only isolated aspects of the problem, limiting their effectiveness in handling complex multi-view scenarios.

Despite their potential, existing 3DVG models struggle to disentangle targets from anchors in multi-anchor textual descriptions~\cite{chang2024mikasa,geng2024viewinfer3d,man2024situational}. Large language models (LLMs) often have difficulty interpreting such descriptions~\cite{hong20233d, yuan2024visual}, yet resolving these ambiguities is crucial for improving grounding accuracy~\cite{huang2024nan}. Compounding this challenge, inconsistencies between textual descriptions and spatial relationships arise when viewpoints change. As illustrated in Fig.~\ref{fig1}, an object described as being to the right of another—such as \textit{``The nightstand is to the right of the bed”}—from a front-facing view may appear on the left when observed from the opposite direction. These perspective-induced inconsistencies make it significantly harder for models to establish accurate correspondences between textual descriptions and visual information, further degrading performance.
Ultimately, both the inherent complexity of multi-anchor queries and the challenges introduced by perspective shifts hinder the accurate interpretation of positional relationships in 3DVG, limiting the overall effectiveness of existing systems.

To tackle these challenges, we propose \textit{ViewSRD}, a framework that formulates 3D visual grounding as a structured multi-view decomposition process. By leveraging the \textit{Simple Relation Decoupling (SRD)} module, ViewSRD effectively disentangles target-anchor relationships in the complex multi-anchor queries, while the \textit{Multi-view Textual-Scene Interaction (Multi-TSI)} module integrates multi-view information to enhance grounding accuracy.
As illustrated in Fig.~\ref{fig1}(b), ViewSRD first applies the SRD module to decompose complex multi-anchor queries into a set of simpler single-anchor queries, isolating interactions between the target and its anchors. This structured decomposition allows the model to more effectively learn positional relationships from textual descriptions. The Multi-TSI module then fuses textual and scene features across multiple viewpoints using \textit{Cross-modal Consistent View Tokens (CCVTs)}, which explicitly encode viewpoint information as learnable cue for both textual and scene module. This mechanism ensures that the model accurately captures spatial interactions, even under perspective shifts.
Finally, the \textit{Textual-Scene Reasoning} module aggregates these multi-view features to accurately predict the final 3D VG results. Extensive experiments have validated the efficacy of our proposed ViewSRD across different 3DVG benchmarks, demonstrating its superior performance across diverse scenarios.
In summary, our contributions are fourfold:
\begin{itemize}
    \item We propose ViewSRD, a framework that formulates 3D visual grounding as a structured multi-view decomposition process, effectively handling complex multi-anchor queries and mitigating text-visual inconsistencies across different perspectives.
    \item We introduce the Simple Relation Decoupling (SRD) module, which restructures complex multi-anchor queries into simpler single-anchor statements, disentangling target-anchor relationships. This structured decomposition enables the model to extract more effective textual features for grounding.
    \item We develop the Multi-view Textual-Scene Interaction (Multi-TSI) module to explicitly encode viewpoint information using cross-modal consistent view tokens. This mechanism ensures alignment between textual descriptions and visual features across different perspectives, reducing spatial ambiguities.
    \item We conduct extensive evaluations on 3D visual grounding datasets, where ViewSRD achieves state-of-the-art performance, yielding superior performance over prior work.
\end{itemize}

\section{Related Work}
\textbf{3D Visual Grounding.}
3D computer vision has made great progress in various fields~\cite{zheng2025recdreamer,pan2025gaussian,liu2025rotation,xu2024learning,zheng2024learning,lin2025delving,chen2025scjd}, the 3D visual grounding (3DVG) task involves identifying a target object in a 3D scene based on a natural language description~\cite{xu2024multi, huang2024structure}. Pioneering datasets such as ScanRefer~\cite{chen2020scanrefer} and ReferIt3D~\cite{achlioptas2020referit3d}, built on ScanNet~\cite{dai2017scannet}, have driven progress in this field. Recent advancements like MVT~\cite{huang2022multi} address view inconsistency by developing a view-robust multi-modal representation. Other works~\cite{zhang2024towards,linghu2025multi} explore multi-modal situated reasoning but lack a dedicated focus on handling the high semantic complexity of natural language in 3D grounding, particularly in disentangling intricate sentence structures.

Despite these advancements, the complexity of natural language descriptions remains a significant challenge in grounding tasks. Referring expressions often require reasoning over multiple anchor objects to precisely identify the target, making it crucial to disentangle and interpret intricate linguistic structures and spatial dependencies. Our method addresses this challenge by decoupling complex queries into simpler statements, improving the extraction of key relational information. Additionally, by leveraging view tokens, ViewSRD learns more accurate associations between textual descriptions and multi-view information.

\begin{figure*}[ht]
    \centering
    \includegraphics[width=1\linewidth]{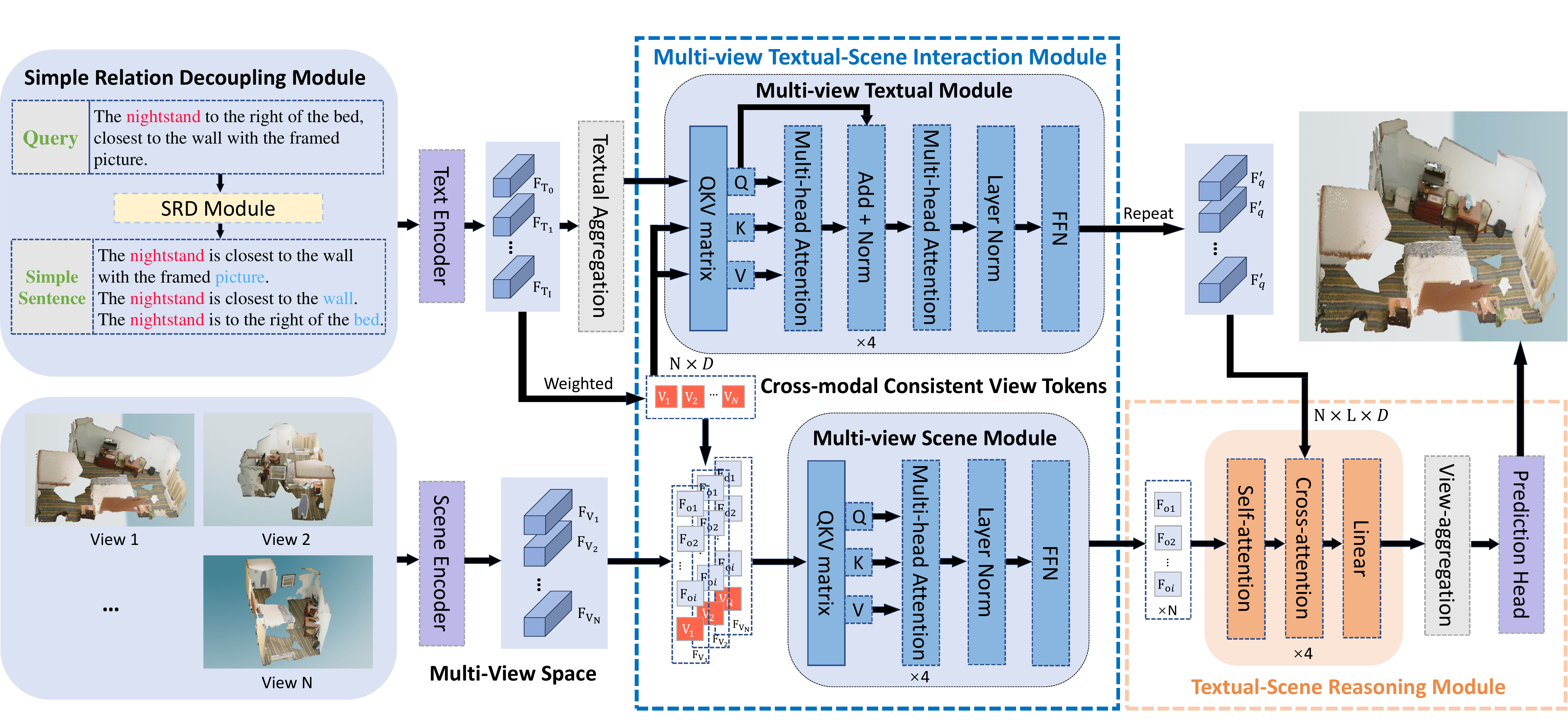}
    \caption{Overview of \textit{ViewSRD}. We begin by employing the \textit{Simple Relation Decoupling (SRD)} module to decompose complex multi-anchor queries into multiple simpler single-anchor queries. Next, text and scene features are extracted separately using the text encoder and scene encoder. To explicitly incorporate scene information into the model, we fuse \textit{Cross-modal Consistent View Tokens (CCVTs)} with these extracted features. The \textit{Multi-view Textual-Scene Interaction (Multi-TSI)} module then facilitates a comprehensive interaction between textual and scene information, the 3DVG prediction results are finally generated by the \textit{Textual-Scene Reasoning Module}.}
    \label{fig:2}
    \vspace{-10pt}
\end{figure*}

\noindent\textbf{Language Comprehension.}
Understanding referential language in 3DVG requires models to not only parse spatial descriptions but also interpret object relationships within a scene. Scene graphs, where objects serve as nodes and relationships form directed edges, have been widely used for tasks such as image retrieval and caption evaluation~\cite{huang2024structure}. Traditional approaches employ scene graphs to enhance query comprehension~\cite{wu2023eda}, with efforts to convert sentences into structured representations~\cite{farshad2023scenegenie, phueaksri2023approach} or generate grounded scene graphs for images~\cite{kundu2023ggt,yu2023visually}. However, these methods primarily focus on static, well-defined relationships and struggle with the dynamic, context-dependent nature of natural language. In datasets such as Nr3D~\cite{achlioptas2020referit3d}, the complexity of interwoven spatial relationships and ambiguous references makes direct scene graph construction challenging. To address this, we propose leveraging Large Language Models (LLMs)~\cite{wang2023openchat,liu2024deepseek,yang2024qwen2} to enhance semantic understanding and spatial reasoning, reducing reliance on rigid structures while improving language comprehension.

\noindent\textbf{3D Multi-View Learning.}
3D vision research has largely focused on generating 2D projections from multiple viewpoints. While LLM-based grounding methods integrate multi-view images, they struggle with accurately identifying the primary viewpoint and demonstrating reliability, as discussed in~\cite{hong20233d,yuan2024visual}. MVT~\cite{huang2022multi} maps 3D scenes into multiple perspectives to enhance cross-view feature aggregation but lacks a mechanism to weigh each view’s contribution, limiting performance in complex scenes. Similarly, ViewRefer~\cite{guo2023viewrefer} utilizes multi-view prototypes for cross-view interactions but lacks explicit training guidance on view importance. Mikasa~\cite{chang2024mikasa} incorporates relative spatial coordinate information and a scene-aware module to improve object grounding but does not fully resolve view weighting challenges. In contrast, we propose Cross-modal Consistent View Tokens, which guide the model to dynamically adjust representation spaces and assess whether spatial relationships in decoupled sentences exhibit view dependency. This mechanism enables more reliable multi-view reasoning, improving performance in complex scenes.

\section{ViewSRD}
% \section{ViewSRD}

% 一、任务背景与挑战
In the context of 3D point cloud scenes, the term \textit{multi-view} refers to observing a shared scene representation (e.g., XYZ+RGB format) from different simulated viewpoints by rotating the scene around its central axis or camera viewpoints. Each view provides a partial observation of the same 3D environment, resulting in varying object appearances, occlusions, and spatial configurations across views. This multi-view setup introduces significant challenges for 3D visual grounding: (1) language-grounded spatial relations must remain consistent across view-dependent variations, and (2) object referents may be partially or completely invisible in certain views. 

% 二、方法设计动机与框架简介
To tackle these challenges, we propose {ViewSRD}, a structured multi-view 3D visual grounding framework. The overall framework of our method is illustrated in Fig.~\ref{fig:2}. ViewSRD comprises two key components. 
The first component is the \textit{Simple Relation Decoupling (SRD)} module, which decomposes multi-anchor queries into a series of single-anchor queries by leveraging the powerful language processing capabilities of LLMs and predefined prompt templates. This decomposition enables more precise inference of relative relationships between objects, improving the model’s ability to capture spatial interactions. 
The second is the \textit{Multi-view Textual-Scene Interaction (Multi-TSI)} module, which mitigates viewpoint dependency by integrating a shared, cross-modal consistent view token into both the language and visual models. These tokens facilitate feature interaction across perspectives, allowing the visual and textual models to align cross-modal viewpoint information more effectively.

\subsection{Simple Relation Decoupling Module\label{sec:SRD}}
\begin{figure}[t]
    \centering
    \includegraphics[width=1\linewidth]{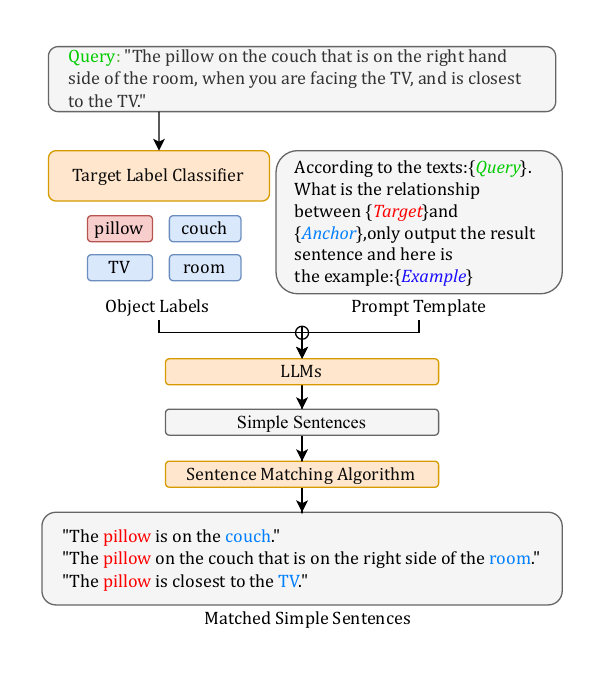}
    \caption{Overview of the SRD Module pipeline. }
    \label{fig4}
    \vspace{-0.5cm}
\end{figure}

The \textit{Simple Relation Decoupling (SRD)} module is designed to structurally decompose a multi-anchor query into multiple simpler single-anchor queries, enhancing the text encoder’s ability to comprehend and process relational information.
As illustrated in Fig.~\ref{fig4}, the SRD module first predicts the target and anchor labels within a sentence, assigning them as the subject and object in the simplified sentence, respectively. This restructuring forms the foundation for generating a structured prompt, which is then fed into an LLM to produce a set of simplified queries.
To maintain semantic integrity, we employ a Sentence Matching algorithm which is described in detail in the supplementary material. that filters and retains the most relevant simplified queries, ensuring that the refined queries faithfully preserve the original meaning while improving clarity and interpretability. By disentangling object relationships between the target and multiple anchors, the SRD module enables more precise relational reasoning, enhancing 3DVG performance.

\noindent\textbf{Target and anchors digging. }
We pre-train a classifier \( Clas \) to identify the word in a sentence that corresponds to the \{\textcolor{red}{\textit{Target}}\} object. Given an input sentence, \( Clas \) first determines which word belongs to the target. Subsequently, we assess whether other words in the sentence appear in the predefined anchor set, \( A_{lab} = \{A_{lab1}, A_{lab2}, ...\} \), provided by the dataset. If a word matches an entry in this set, it is classified as an \{\textcolor{cyan}{\textit{Anchor}}\} object. For more details about the \( Clas \), please refer to the \textit{supplementary materials}.

\noindent\textbf{Decoupled multi-anchor queries. } 
In practical referring queries, multiple anchors frequently co-occur within the same sentence, and the spatial relationship of the target is inherently tied to the anchor labels. In such cases, spatial descriptions involving multiple objects and their attributes often become entangled. For instance, in the \{\textcolor{green}{\textit{Query}}\} illustrated in Fig.~\ref{fig4}, the object ``couch", which is near the target ``pillow", may dominate the spatial description, thereby weakening the relationship between the target and other anchors.  
To address the coupling issue in such queries, we design a set of prompt templates based on prior target and anchor digging, the process is shown in Fig.~\ref{fig4}. Leveraging the reasoning capabilities of LLMs, we decompose complex multi-anchor queries into simpler single-anchor queries. This decoupling process clarifies the positional relationships between the target and its anchors in 3DVG, enhancing the model’s spatial understanding.  

We define a total of $k$ structured \{\textcolor{blue}{\textit{Example}}\} derived from our pre-designed templates, such as \textit{``The target is on the anchor”}, with additional examples provided in the supplementary materials. For each anchor, the model generates \( k \) candidate queries, where \( k \) denotes the number of generated examples. To ensure the selected sentence best aligns with the original query, we apply a sentence-matching algorithm that evaluates both label consistency and semantic consistency. The final ranking is determined by a weighted average of these two scores. For further details, please refer to the \textit{supplementary materials}.

\subsection{Textual Aggregation}\label{sec:3.2}
Given a complex multi-anchor query sentence, the SRD module decomposes it into \( (I+1) \) sentences, where \( I \) represents the number of anchors in the original sentence. Each anchor contributes to a shorter, simplified sentence, while the original complex query remains as a longer reference sentence. To extract meaningful linguistic representations, we employ BERT~\cite{devlin2018bert} as the text encoder to extract \( (I+1) \) sentence features, generating a language feature set \( \{F_{T_0}, F_{T_1}, \dots, F_{T_I}\} \), where \( F_{T_0} \) corresponds to the original complex query, and the remaining elements represent the features of the decoupled simpler queries.  
To enable the model to effectively learn from diverse sentence representations, we introduce a textual feature aggregation strategy. We randomly sample one feature from the language feature set as the main feature \( F_{\text{main}} \), while treating the remaining features as auxiliary features \( F_{\text{aux}} \). The final aggregated feature is computed as:  
\begin{equation}
  F_{\text{agg}} = \alpha F_{\text{main}} + (1 - \alpha) \cdot \frac{1}{I} \sum_{i=1}^{I} F_{\text{aux}_{i}},  
\end{equation}
where \( \alpha \) is uniformly sampled from \( \{0, 0.1, 0.3, 0.5\} \) during training and fixed at 0.5 during validation. This adaptive fusion strategy ensures smooth feature integration, enhancing the model’s robustness in language-conditioned 3DVG.

\subsection{Multi-view Textual-Scene Interaction Module\label{sec:vt}}
\noindent\textbf{Cross-modal Consistent View Tokens.}
Previous methods have largely overlooked the inconsistency in textual descriptions arising from perspective shifts in multi-view VG, making it challenging for models to accurately interpret these variations \cite{beckham2023visual,wang2024g}. To address this limitation, we introduce a series of learnable and shared \textit{Cross-modal Consistent View Tokens(CCVTs)}, which are integrated into both the textual and scene modules. By incorporating these tokens, both models are explicitly guided with relevant perspective information, enabling them to more effectively capture and understand the transformations and interactions induced by viewpoint changes.  

Formally, let \( \mathcal{V} = \{ V_n | n=1,2,\dots,N; V_n \in \mathbb{R}^D \} \) represents the set of CCVTs, where \( N \) denotes the number of viewpoints and \( D \) represents the dimensionality of CCVTs. The CCVTs are jointly optimized with our proposed textual and scene modules. Once trained, their values remain fixed during inference, serving as a stable reference that enhances the model’s ability to comprehend multi-view scenarios and resolve perspective-induced inconsistencies.

\noindent\textbf{Multi-view Textual Module.} 
To effectively integrate sentence features from text encoders with viewpoint features extracted from CCVTs, we introduce the \textit{Multi-view Textual Module}, which employs a cross-attention mechanism \cite{vaswani2017attention} to seamlessly encode viewpoint features \( \mathcal{V} \) into the textual feature space through multi-head attention operation.

Since each sentence inherently carries distinct viewpoint information, it is crucial to embed perspective-aware features into textual representations effectively. To achieve this, we first compute the normalized dot product between each view token and the 0th token of each sentence’s language feature \( \{F_{T_0}, F_{T_1}, \dots, F_{T_I}\} \), as the 0th token \( F^0 \) typically aggregates the most salient semantic information. We take the average of these dot products across different sentences and compute a corresponding probability distribution using the softmax function. This probability is then used to reweight the view token, adaptively increasing its contribution when the description aligns with the viewpoint and reducing it when the description does not match. The refined viewpoint token is formulated as:
\begin{equation}
\mathcal{V} = \text{Softmax} \left(\frac{1}{I} \sum_{i=0}^{I} \frac{F_{T_i}^0 \mathcal{V}^T} {\|F_{T_i}^0\| \cdot \|\mathcal{V}\|} \right) \mathcal{V}.
\end{equation}

Subsequently, the aggregated features \( F_{agg} \), as introduced in Section~\ref{sec:3.2}, serve as the query, while the viewpoint features \( \mathcal{V} \) act as the key and value in the attention computation. The textual feature enriched with viewpoint embeddings, denoted as \( F_q' \), is formulated as:  
\begin{equation}\label{eq:new1}
\begin{aligned}
F_q' = \text{Softmax}\left(\frac{(W_q F_{agg})(W_k \mathcal{V})^T}{\sqrt{D}}\right) W_v \mathcal{V},
\end{aligned}
\end{equation}  
where \( W_q \), \( W_k \), and \( W_v \) are learnable linear projection matrices. Following this, \( F_q' \) undergoes an additional self-attention operation to further refine the textual features, ensuring that the encoded representations effectively capture perspective-dependent information.

\noindent\textbf{Multi-view Scene Module.}
To effectively capture object features across diverse scenes, we introduce a \textit{Multi-View Scene Module} that extracts and refines scene representations from multiple viewpoints. To achieve this, we employ PointNet++ \cite{qi2017pointnet++} as the scene encoder, computing scene features \( F_{V_n} \) for each viewpoint, where \( n \in N \) denotes the scene index across \( N \) viewpoints. Each scene feature \( F_{V_n} \) consists of object-level representations, expressed as \( \{F_{o1}, F_{o2}, \ldots, F_{oi}\} \), where \( i \) corresponds to the number of objects present in the scene. 

To explicitly inform the model of the current scene, we concatenate our CCVTs \( V_n \) with the extracted scene features, forming the input representation:  
\begin{equation}\label{eq:new2}
\begin{aligned}
\mathbf{X}_n = \{ F_{V_n}, V_n \} = \{ F_{o1}, F_{o2}, \ldots, F_{oi}, V_n \}.
\end{aligned}
\end{equation}
These combined feature representations are then processed through several Transformer layers \cite{vaswani2017attention}, denoted as $\mathrm{Trans}(\cdot)$, which enhances the relational encoding between objects and viewpoints. This mechanism ensures that both global scene context and fine-grained object details are effectively captured:  
\begin{equation}\label{eq:new3}
\begin{aligned}
\mathbf{Z}_n^{(l+1)} = \text{Trans}^{(l)}(\mathbf{Z}_n^{(l)}),
\end{aligned}
\end{equation} 
where the initial input to the Transformer is \( \mathbf{Z}_n^{(0)} = \mathbf{X}_n \), and \( \mathbf{Z}_n^{(L)} \) represents the refined features after \( L \) Transformer layers.  

At the final Transformer layer, the output consists of both [\textit{object}] tokens and [\textit{view}] tokens. Since the transformed features \( {F}_{V_n}' \) encapsulate both object-specific and viewpoint information, we retain only the [\textit{object}] tokens for the subsequent grounding task: \begin{equation}\label{eq:new4}
\begin{aligned}
{F}_{V_n}' = \{ {F}_{o1}', {F}_{o2}', \ldots, {F}_{oi}' \}.
\end{aligned}
\end{equation} 
This design ensures that object representations are enriched with multi-view contextual information while maintaining their distinct semantic properties for accurate 3DVG.

\begin{figure*}[ht!]
    \centering
    \includegraphics[width=\linewidth]{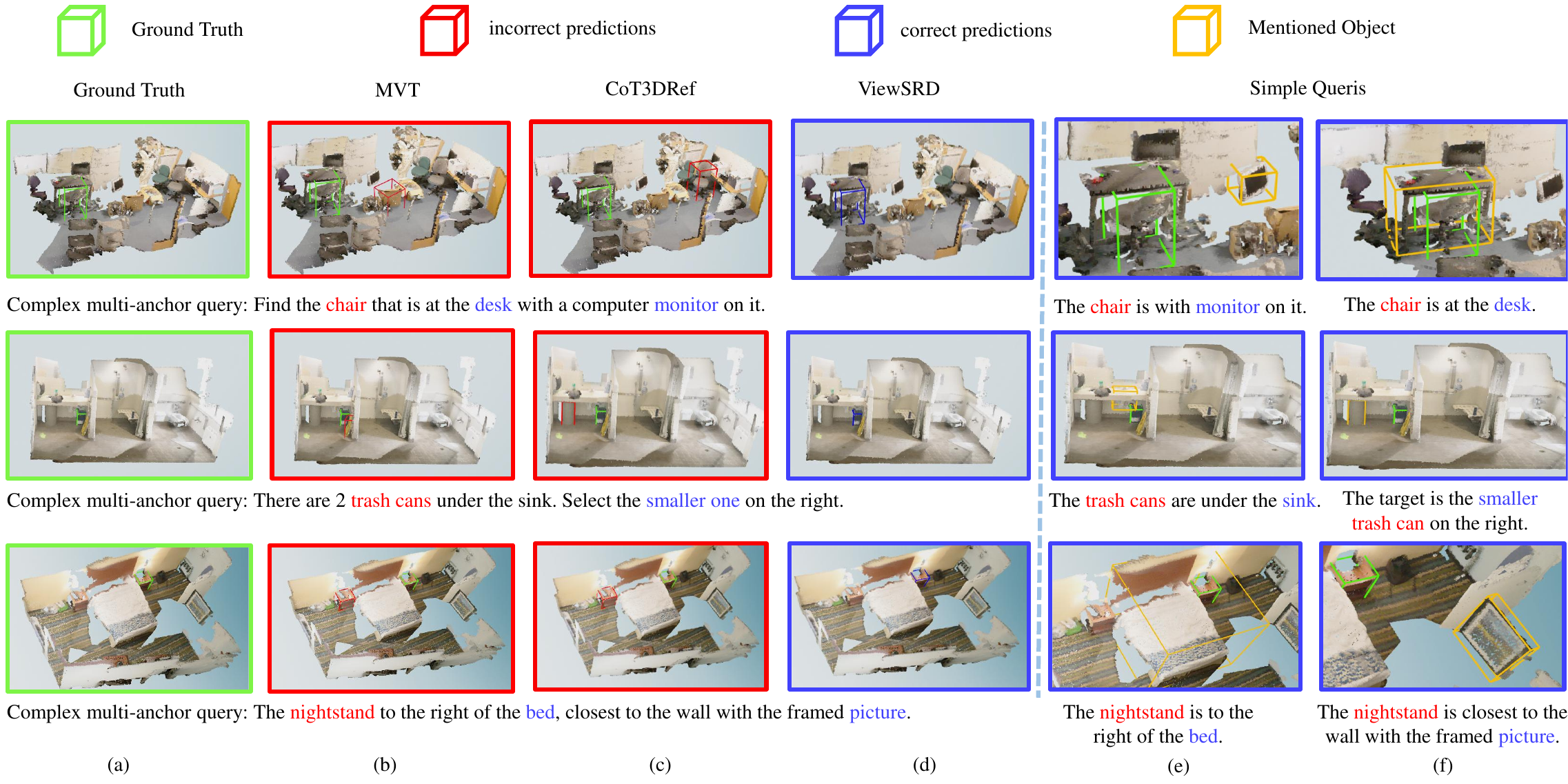}
    \caption{Visualization Results of the 3D Visual Grounding Results.  For the presented 3D scenes, we utilize green, red, blue, and yellow boxes to represent the ground truth, incorrect predictions, correct predictions, and the mentioned objects, respectively. Columns (e) and (f) present the decomposed simple queries derived from the complex querys.}
    \label{fig5}
    \vspace{-15pt}
\end{figure*}

\subsection{Textual-Scene Reasoning Module}
With the above formulation, we obtain the view-interactive textual features \( F_q' \) and scene features \( F_{V}' \), where \( F_{V}' = \{F_{V_n}' \mid n = 1,2,\dots,N\} \), each enriched with viewpoint information. These features are then processed through the proposed \textit{Textual-Scene Reasoning Module} to generate the final prediction.  
This module primarily consists of a Transformer with a cross-attention mechanism~\cite{vaswani2017attention}, where \( F_{V}' \) serves as the query, while \( F_q' \) functions as the key and value, facilitating fine-grained alignment between textual and visual representations. Additionally, a View Aggregation mechanism integrates information across multiple viewpoints by computing both the average and maximum values of the output features. Finally, a Prediction Head projects the aggregated features into the result space, enabling a view-aware 3D Visual Grounding model capable of effectively reasoning across multiple perspectives.

\subsection{Overall Loss Functions}\label{sec:3.5}
Following prior research \cite{roh2022languagerefer,huang2022multi}, three distinct loss functions are applied on \textit{ViewSRD}. These include a referential loss \(\mathcal{L}_{Ref}\) derived from grounding predictions, an object-level loss \(\mathcal{L}_{Object}\) capturing object shape and center and a sentence-level loss \(\mathcal{L}_{Sent}\) designed to identify the target and anchor phrases within the \(F_{\text{agg}}\). Similarly to \cite{bakr2024cot3dref}, we extend the referential loss  \(\mathcal{L}_{Ref}\) to localize both the target and the anchors, which we term as parallel referential loss \( \mathcal{L}^{P}_{\text{ref}} \), where both the target and anchors are localized simultaneously. For details of these losses, please refer to \textit{supplementary materials}. The total loss function is defined as:
\begin{equation}\label{eq:w8}
\begin{aligned}
\mathcal{L} =  \lambda_{Obj}\mathcal{L}_{Object}+\lambda_{Ref} \mathcal{L}^{P}_{Ref}+\lambda_{Sent}\mathcal{L}_{Sent}.
\end{aligned}
\end{equation}

\section{Experiments}

\subsection{Experiment Settings}

\noindent\textbf{Datasets.} \textbf{Nr3D}~\cite{achlioptas2020referit3d} contains 45,503 human utterances referencing 707 indoor scenes from ScanNet~\cite{dai2017scannet}, covering 76 object categories with multiple same-class distractors. \textbf{Sr3D}~\cite{achlioptas2020referit3d} includes 83,572 template-based sentences in a ``target–spatial relation–anchor'' format, offering a simpler setup with similar distractors. \textbf{ScanRefer}~\cite{chen2020scanrefer} provides 51,583 free-form descriptions for 11,046 objects across 800 ScanNet scenes, incorporating spatial and attribute-level references to support 3DVG.

\noindent\textbf{Evaluation Metrics.} For Nr3D and Sr3D, grounding accuracy is measured by the percentage of correctly matched boxes~\cite{huang2022multi, roh2022languagerefer}. For ScanRefer, we report Acc@0.25 and Acc@0.5, i.e., the percentage of predicted boxes with IoU exceeding 0.25 or 0.5, respectively~\cite{wu2023eda}.  

% \noindent\textbf{Loss Weight Settings.}
% To ensure robustness across different datasets, we conduct a systematic ablation study on the loss weight coefficients $\lambda_{Ref}$, $\lambda_{Obj}$, and $\lambda_{Sent}$. The configuration $(1.0, 0.5, 0.5)$ yields the highest grounding accuracy while maintaining stability. Please refer to the supplementary material (Table 1) for the full results and discussion.

\noindent\textbf{Implementation Details.} All experiments are implemented in PyTorch and run on a single RTX 4090 GPU. We use AdamW~\cite{loshchilov2017decoupled,yang2024g,kang2025sita,xu2022multi} with a learning rate of 0.0005. The number of input views is set to $N = 4$. We set $\lambda_{Ref},\lambda_{Obj},\lambda_{Sent} = 1.0, 0.5, 0.5$. For the SRD module, we adopt DeepSeek-R1~\cite{liu2024deepseek}, which balances performance and reproducibility, and can runs on a RTX 4090.

\subsection{3D Visual Grounding Results}
\begin{table*}[h]
\centering
\caption{Performance (\%) comparison on Nr3D \cite{achlioptas2020referit3d} and Sr3D  \cite{achlioptas2020referit3d}.}
\vspace{-10pt}
\resizebox{1.0\textwidth}{!}{
\begin{tabular}{lcccccccccc}
\toprule
\multirow{2}{*}{Method} & \multicolumn{5}{c}{Nr3D}  & \multicolumn{5}{c}{Sr3D}         \\ \cmidrule(l){2-11} 
 &
  Overall &
  Easy &
  Hard &
 View Dep. &
  View Indep. &
  Overall &
  Easy &
  Hard &
  View Dep.&
  View Indep. \\ \midrule

% ReferIt3D \cite{achlioptas2020referit3d}    & 35.6 & 43.6 & 27.9 & 32.5 & 37.1 & 40.8  & 44.7  & 31.5  & 39.2  & 40.8 \\
% InstanceRefer \cite{yuan2021instancerefer}    & 38.8 & 46.0 & 31.8 & 34.5 & 41.9 & 48.0  & 51.1  & 40.5  & 45.4  & 48.1 \\
3DVG-Transformer \cite{zhao20213dvg}   & 40.8 & 48.5 &34.8 & 34.8 & 43.7 & 51.4  & 54.2  & 44.9  & 44.6  & 51.7  \\
LanguageRefer \cite{roh2022languagerefer}    & 43.9 & 51.0 & 36.6 & 41.7 & 45.0 & 56.0  & 58.9  & 49.3  & 49.2  & 56.3 \\
TransRefer3D \cite{he2021transrefer3d}   & 42.1 & 48.5 & 36.0 & 36.5 & 44.9 & 57.4  & 60.5  & 50.2  & 49.9  & 57.7  \\
SAT \cite{yang2021sat}        & 49.2 & 56.3 &42.4 & 46.9 & 50.4  & 57.9  & 61.2  & 50.0  & 49.2  & 58.3 \\
MVT   \cite{huang2022multi}      & 55.1  & 61.3  & 49.1  & 54.3  & 55.4 & 64.5 & 66.9 & 58.8 & 58.4 & 64.7  \\
ViewRefer \cite{guo2023viewrefer}  & 56.0  & 63.0  & 49.7  & 55.1  & 56.8 & {67.0} & {68.9} & {62.1} & 52.2 & {67.7}  \\ 
MiKASA \cite{chang2024mikasa}  & 64.4  & 69.7  & 59.4  & 65.4  & 64.0 & {75.2} & {\textbf{78.6}} & {67.3} & \textbf{70.4} & {75.4}  \\ 
CoT3DRef \cite{bakr2024cot3dref}  & 64.4  & 70.0  & 59.2  & 61.9  & 65.7 & {73.2} & {75.2} & {67.9} & 67.6 & {73.5}  \\ 
% \midrule
% MVT*  \cite{huang2022multi}       & 54.5  & 61.2  & 48.0  & 53.0  & 55.2 & 64.2 & 67.3 & 57.0 & 55.6 & 64.6  \\
ViewSRD (ours) & \textbf{69.9}  & \textbf{75.3}  & \textbf{64.8}  & \textbf{68.6}  & \textbf{70.6} & \textbf{76.0} & 78.3     & \textbf{70.6}     & {69.0 }   &  \textbf{76.2} \\

\bottomrule

\end{tabular}
}
\label{Table 1:}
\vspace{-10pt}
\end{table*}

We compare \textit{ViewSRD} with recent state-of-the-art approaches to evaluate its effectiveness on 3DVG.
Fig.~\ref{fig5} illustrates complex query cases from Nr3D~\cite{achlioptas2020referit3d}, including ground truth boxes, predictions from MVT~\cite{huang2022multi}, CoT3DRef~\cite{bakr2024cot3dref}, and \textit{ViewSRD}, along with the original queries and the simplified sentences produced by the SRD module. In multi-anchor scenarios (e.g., involving “bed”, “table”, and “chair”), MVT and CoT3DRef often misalign predictions due to challenges in spatial reasoning. In contrast, \textit{ViewSRD} correctly grounds targets by decomposing complex queries and leveraging robust spatial relationships between target-anchor pairs. Moreover, under viewpoint shifts, CoT3DRef struggles to maintain alignment, whereas \textit{ViewSRD} reliably grounds targets by capturing spatial relations invariant to viewpoint changes (e.g., “The trash cans are under the sink”).

Quantitative results on Nr3D (Table~\ref{Table 1:}) show that \textit{ViewSRD} achieves a 5.2\% accuracy gain over the best prior method, CoT3DRef, under identical settings. Under view-dependent evaluation, it further outperforms CoT3DRef by 6.7\%, demonstrating the effectiveness of CCVTs in aligning textual and visual spaces and modeling viewpoint-sensitive relations through query decomposition.
To assess generalization, we also evaluate on Sr3D~\cite{achlioptas2020referit3d} (Table~\ref{Table 1:}). \textit{ViewSRD} achieves the highest accuracy of 76.2\% in the View-Independent setting, with additional gains of +2.8\% and +2.7\% in the View-Independence and Hard scenarios, respectively. These results confirm the robustness and generalizability of our approach across diverse scenario.

\subsection{Analysis of Anchors}
\label{sec4}
In this section, we analyze the impact of the number of anchors in a query on 3DVG performance. The results presented in Table \ref{tab2} underscore the effectiveness of our approach, particularly in multi-anchor scenarios, where our method successfully disentangles spatial relationships by explicitly modeling target-anchor interactions.
In contrast, existing methods such as MVT~\cite{huang2022multi} and CoT3DRef~\cite{bakr2024cot3dref}, which do not account for the necessity of spatial relationship decoupling, exhibit a notable performance decline in multi-anchor queries compared to single-anchor cases. Notably, our approach achieves higher accuracy in multi-anchor queries than in single-anchor ones, demonstrating that when properly processed, multi-anchor information enhances 3DVG performance rather than introducing ambiguity. These findings validate the efficacy of ViewSRD in effectively leveraging complex spatial relationships for improved grounding accuracy.
\begin{table}[h]
\centering
  \caption{Performance (\%) comparison on Nr3D~\cite{achlioptas2020referit3d} with new criterions Multi-Anc and Single-Anc.}
  \vspace{-10pt}
  \label{tab2}
  \small
  \begin{tabular}{lccc}
    \toprule
    Model & Multi-Anc & Single-Anc & Overall \\
    \midrule
    {MVT \cite{huang2022multi}} & 52.6 & 56.6 & 55.1 \\
    {CoT3DRef \cite{bakr2024cot3dref}} & 63.1 & 65.2& 64.4 \\
    {ViewSRD} & \textbf{71.5} & \textbf{69.5}& \textbf{69.9} \\
    \bottomrule
  \end{tabular}
\end{table}

\subsection{SRD Enhances Other 3DVG Methods.}
\begin{table}[t]
    \centering
    \caption{Performance (\%) of SRD module improves MVT \cite{huang2022multi}, BUTD-DETR \cite{jain2022bottom} and EDA \cite{wu2023eda} on ScanRefer \cite{chen2020scanrefer} dataset.}
    \vspace{-10pt}
    \label{tab:3D_VG_results}
    \renewcommand{\arraystretch}{1.2}
    \resizebox{\linewidth}{!}{
    \begin{tabular}{l c c c c c c}
        \toprule
        \multirow{2}{*}{Method} & \multicolumn{2}{c}{Unique (19\%)} & \multicolumn{2}{c}{Multiple (81\%)} & \multicolumn{2}{c}{Overall} \\
        \cmidrule(lr){2-7}
        & 0.25 & 0.5 & 0.25 & 0.5 & 0.25 & 0.5 \\
        \midrule
        MVT \cite{huang2022multi} & 77.7 & 66.5 & 31.9 & 25.3 & 40.8 & 33.3 \\
        
        MVT+SRD & 78.6 & 67.2 \ & 34.1 & 27.1 & 42.1 (\red{3.2\%↑}) & 34.3(\red{3.0\%↑}) \\
        \midrule
        BUTD-DETR \cite{jain2022bottom} & 82.8 & 64.9 & 44.7 & 33.9 & 50.4 & 38.6 \\
        
         BUTD-DETR+SRD & {85.0} & 66.2 & {45.3} & {34.2} & {57.9} (\red{14.9\%↑}) & {45.7} (\red{18.4\%↑}) \\
         \midrule
        EDA \cite{wu2023eda} & 80.4 & 65.3 & 35.6 & 25.1 & 43.6 & 32.3 \\
        
         EDA+SRD & {81.0} & 67.3 & {36.4} & {28.3} & {44.4} (\red{1.8\%↑}) & {35.3} (\red{9.3\%↑}) \\
         \midrule
         ViewSRD & 82.1 & 68.2 & 37.4 & 29.0 & 45.4 & 36.0 \\
         
        \bottomrule
    \end{tabular}
    }
    \vspace{-10pt}
\end{table}
Our SRD module is inherently model-agnostic, operating independently of the training process by focusing exclusively on decoupling complex multi-anchor queries into simpler single-anchor queries. This decoupling mechanism reduces ambiguity in multi-anchor descriptions, enhances target grounding, and serves as a model-independent preprocessing step, ensuring seamless compatibility with various 3DVG methods to improve performance without modifying existing architectures.
As demonstrated in Table~\ref{tab:3D_VG_results}, integrating SRD into MVT \cite{huang2022multi}, BUTD-DETR \cite{jain2022bottom} and EDA \cite{wu2023eda} consistently leads to performance improvements. These results highlight SRD’s ability to refine query interpretation by effectively disentangling target-anchor relationships, thereby reducing errors introduced by complex linguistic structures. These improvements reinforce the critical role of SRD module in enhancing accuracy of 3DVG.

% rebuttal Gs8n#Q3  把EDA加了上去：and EDA \cite{wu2023eda}

\subsection{Ablation Study}
\begin{table}[t]
\centering
\caption{Ablation studies on Nr3D~\cite{achlioptas2020referit3d}. All components contribute to final performance(\%).}\label{Table al}
\vspace{-10pt}
\small
\setlength{\tabcolsep}{4pt}
\begin{tabular}{@{}lccccc@{}}
\toprule
Component               & Overall & Easy & Hard & View Dep. & View Indep.  \\ \midrule
w/o CCVTs.     & 62.2       & 68.5    & 56.1    & 60.1  & 63.2   \\
w/o Textual M.     & 68.0        & 73.5  & 62.6    & 67.6  & 68.1   \\
w/o Scene M.     & 64.6       & 70.5    & 58.9    & 63.8  & 64.9  \\
w/o SRD M.     & 68.6      & 73.0    & 64.8     & 66.5  & 70.0  \\ 
w/o Weight.      &  69.0  &  74.2  &  64.0  &  66.5 & 70.2 \\
LLM-Aug.      &  69.1  &  74.5  &  63.7  &  68.0 & 69.5 \\
ViewSRD & \textbf{69.9}  & \textbf{75.3}  & \textbf{64.8}  & \textbf{68.6}  & \textbf{70.6} \\
\bottomrule
\end{tabular}
\vspace{-10pt}
\end{table}
\begin{table}[t]
    \centering
    \caption{Ablation of view numbers on Nr3D~\cite{achlioptas2020referit3d}.}
    \vspace{-10pt}
    \label{tab:view_numbers}
    \renewcommand{\arraystretch}{1.1} % 调整行距
    \setlength{\tabcolsep}{4pt} % 调整列间距
    \small 
    \resizebox{\linewidth}{!}{
    \begin{tabular}{cc|ccccc}
        \toprule
        \multicolumn{2}{c|}{View Number} & \multirow{2}{*}{Overall} & \multirow{2}{*}{Easy} & \multirow{2}{*}{Hard} & \multirow{2}{*}{View Dep.} & \multirow{2}{*}{View Indep.} \\
        Train & Test &  &  &  &  &  \\
        \midrule
        4 & 1 & 66.0  & 71.7  & 60.5  & 64.0  &67.0  \\
        4 & 2 & 68.9  & 75.1  & 63.0  & 66.9  &69.9  \\
        4 & 4 & \textbf{69.9 }  & \textbf{75.3 }  & \textbf{64.8 }  & \textbf{68.6 }  & \textbf{70.6 } \\
        % 4 & 8 & 54.8  & 61.1  & 48.8  & 54.2  & 55.1  \\
        \midrule
        1 & 1 & 64.4  & 70.9  & 58.1  & 60.8  & 66.2  \\
        2 & 2 & 67.7  & 73.0  & 62.5  & 66.1  & 68.4  \\
        8 & 8 & 68.4  & 74.1  & 63.0  & 67.4  & 68.9  \\
        \bottomrule
    \end{tabular}
   }
\vspace{-10pt}
\end{table}
\noindent\textbf{Analysis of ViewSRD Components.} To assess the contribution of each component within ViewSRD, we conducted detailed ablation studies on the Nr3D dataset \cite{achlioptas2020referit3d}. Starting from the full model, we systematically removed key modules one at a time to evaluate their individual impact. The results, presented in Table \ref{Table al}, demonstrate that each component plays a crucial role in enhancing model performance across different scenarios.
Notably, the removal of the CCVTs leads to the most significant performance degradation. This is primarily because, without the view token, the model lacks explicit viewpoint information, impairing its ability to distinguish between different perspectives. Similarly, removing either the textual module or the scene module results in a noticeable decline, underscoring the necessity of cross-modal interaction. When view-alignment weighting is disabled (\textit{w/o Weight}), performance drops by 0.9\%, showing that dynamic alignment of view features is critical for performance under view-dependent conditions. Removing the SRD module leads to performance degradation, confirming the benefit of multi-anchor query decoupling. We also compare it with an LLM-based augmentation method from Multi3DRefer~\cite{zhang2023multi3drefer} and find that SRD achieves greater gains, highlighting the advantage of structured query decomposition over generic augmentation.

% rebuttal VmM6#Q2 VmM6#Q3  上面的weight 和LLM-Aug修改↓
% 修改了最后最后两句，从Removing the SRD module开始。原文为↓
% Additionally, eliminating the SRD module also reduces the effectiveness, confirming that decoupling multi-anchor sentences significantly improves 3DVG accuracy. We also replace SRD with an LLM-based augmentation method from Multi3DRefer. Although both methods improve grounding, SRD yields better gains, suggesting that structured sub-query decomposition provides more precise grounding cues than generic augmentation.

\noindent\textbf{Analysis of Multi-View Modeling.} We evaluate the effect of varying view counts on 3DVG performance using the Nr3D dataset. As shown in Table~\ref{tab:view_numbers}, testing with more views consistently improves accuracy when the model is trained with four views, highlighting the benefit of aggregating complementary spatial cues from multiple perspectives.
When training and testing with the same number of views, performance improves from 64.4\% (1 view) to 67.7\% (2 views), but plateaus at 68.4\% with 8 views, suggesting diminishing returns. Notably, four views offer a strong trade-off, capturing diverse spatial information with minimal redundancy and maintaining computational efficiency. This also suggests that uniformly attending to many views may dilute focus on key perspectives. Future work will explore adaptive view selection.

% \noindent\textbf{Analysis of Multi-View Modeling.} We further investigate the impact of varying the number of views on 3DVG performance during both training and testing on the Nr3D dataset. As shown in Table~\ref{tab:view_numbers}, increasing the number of views during testing consistently enhances performance when the model is trained with four views. This trend underscores the effectiveness of multi-view aggregation in improving grounding accuracy by leveraging complementary spatial information from multiple perspectives.

% Additionally, performance variations emerge when the number of views remains consistent between training and testing. Training and testing with a single view yield an accuracy of 64.4\%, which increases to 67.7\% with two views. However, further increasing the number of views to eight results in only a marginal gain, reaching 68.4\%. These results indicate that while incorporating additional viewpoints benefits performance, the improvements eventually plateau, suggesting a point of diminishing returns. Notably, using four views provides a well-balanced multi-view representation, effectively capturing diverse spatial information while minimizing redundancy and maintaining fast computation. This suggests that equal attention distribution across many views can dilute focus from crucial perspectives. Future work will explore dynamic view selection strategies.
%rebuttal Gs8n#Q5 This开始的上面最后一句

\begin{table}[t]
    \centering
    \caption{Accuracy comparison when replacing different LLMs in SRD module on Nr3D~\cite{achlioptas2020referit3d}.}
    \vspace{-10pt}
    \label{tab:object_encoding}
    \setlength{\tabcolsep}{8pt} % 调整列间距
    \small  % 调整表格内字体大小
    \begin{tabular}{l c}
        \midrule
        LLM decoupler & {Accuracy} \\
        \midrule
        OpenChat~\cite{wang2023openchat} & 69.6\% \\
        DeepSeek-R1~\cite{liu2024deepseek}  & 69.9\% \\
        Qwen-Plus~\cite{yang2024qwen2} & 70.5\%\\ % 加粗最高准确率
        Qwen-Turbo~\cite{yang2024qwen2} & \textbf{70.7\%} \\ % 加粗最高准确率
        \bottomrule
    \end{tabular}
    \vspace{-10pt}
\end{table}
\noindent\textbf{Analysis of SRD's LLM Decoupler.} In this paper, we employ the open-source DeepSeek-R1~\cite{liu2024deepseek} as the LLM in the SRD module and further investigate the impact of different LLMs on the final performance of 3DVG. As shown in Table~\ref{tab:object_encoding}, different LLM decouplers exhibit varying levels of effectiveness in the sentence decoupling task. Models with stronger decoupling capabilities yield better results.
For instance, OpenChat~\cite{wang2023openchat} and DeepSeek-R1~\cite{liu2024deepseek} achieve accuracies of 69.6\% and 69.9\%, respectively, while models designed with enhanced sentence decoupling capabilities, such as Qwen-Plus~\cite{yang2024qwen2} and Qwen-Turbo~\cite{yang2024qwen2}, achieve 70.5\% and 70.7\%, with Qwen-Turbo demonstrating the highest performance. These results indicate that as an LLM’s ability to disentangle complex sentence structures improves, it becomes more effective at isolating and extracting relevant information, ultimately leading to significant gains in 3DVG accuracy.

\section{Conclusion}
In this paper, we introduce ViewSRD, a framework that disentangles target-anchor relationships via the Simple Relation Decoupling (SRD) module and enhances multi-view understanding through the Multi-view Textual-Scene Interaction (Multi-TSI) module. By decomposing complex multi-anchor queries into simpler single-anchor sentences, SRD clarifies positional relationships, while Multi-TSI integrates textual and scene features across viewpoints using cross-modal consistent view tokens (CCVTs) to capture spatial interactions. Extensive experiments demonstrate ViewSRD’s state-of-the-art performance in 3DVG.

A limitation of ViewSRD is its assumption that complex queries can be fully decomposed without overlapping relationships. While the decomposition into overlapping relations  does not degrade performance, it diminishes the intended benefits of simplification. Future work will explore adaptive query to better preserve contextual dependencies.

\noindent\textbf{Acknowledgements.} \footnotesize{This work is supported by Guangdong Provincial Natural Science Foundation for Outstanding Youth Team Project (No. 2024B1515040010), NSFC Key Project (No. U23A20391), China National Key R\&D Program (Grant No. 2023YFE0202700, 2024YFB4709200), Key-Area Research and Development Program of Guangzhou City (No. 2023B01J0022), Guangdong Natural Science Funds for Distinguished Young Scholars (Grant 2023B1515020097), the National Research Foundation, Singapore under its AI Singapore Programme (AISG Award No.: AISG3-GV-2023-011), the Singapore Ministry of Education AcRF Tier 1 Grant (Grant No.: MSS25C004), and the Lee Kong Chian Fellowships.
}

\clearpage
{
    \small
    \bibliographystyle{ieeenat_fullname}
    \bibliography{main}
}
\cleardoublepage
\appendix

\centerline{\textbf{\Large{Appendix}}}
%%%%%%%%% BODY TEXT - ENTER YOUR RESPONSE BELOW

\section{Summary}
This supplementary material offers detailed technical information on key components of our proposed framework, which were abbreviated in the main text. Specifically, we elaborate on the following: the target classifier $Clas$, the examples in template and the details of the sentence matching algorithm in Section 3.1; details of the loss function in Section 3.5.

\section{Target Classifier  $Clas$}
The target label classifier, denoted as \(Clas\), is designed to regress the target's class label from an input query \(Q\). By determining the category of \(Q\), we then determine which word in \(Q\) belongs to that category to determine which word to use as the \textit{Target}. Essentially, \(Clas\) processes an input sentence by first encoding it via a pre-trained BERT \cite{devlin2018bert} model to obtain a semantic embedding, and then maps this embedding through a multi-layer perceptron (MLP) to yield a probability distribution over \(C\) target classes.

Specifically, given an input query \(Q\), we first compute its embedding:
\begin{equation}
F_Q = \mathrm{BERT}(Q),
\end{equation}
where \(F_Q \in \mathbb{R}^{K \times d}\) represents the BERT output for the query \(Q\), with \(K\) being the number of tokens in \(Q\) and \(d\) denoting the feature dimension. We select the embedding corresponding to the [CLS] token as the aggregate representation \(h\). This \(h\) is then fed into the MLP-based classification head of \(Clas\), implemented as a fully connected layer that produces logits \(z \in \mathbb{R}^{C}\). These logits are subsequently converted into a probability distribution over the \(C\) classes via the softmax function:

\begin{equation}
P(c \mid Q) = \frac{\exp(z_c)}{\sum_{j=1}^{C}\exp(z_j)},
\end{equation}

where \(z_c\) denotes the \(c\)-th element of \(z\).

The classifier \(Clas\) is trained by minimizing the standard cross-entropy loss:

\begin{equation}
\mathcal{L}_{\text{cls}} = -\sum_{c=1}^{C} y_c \log\big(P(c \mid Q)\big),
\end{equation}
where \(y_c\) is the one-hot encoded ground-truth label for class \(c\). During inference, the predicted target label is given by:

\begin{equation}
T_{lab} = \arg\max_{c} \, P(c \mid Q).
\end{equation}

In summary, \(Clas\) effectively bridges the input query and the target object label by leveraging BERT-based encoding followed by MLP-based classification, ensuring robust target label regression in language-conditioned 3D grounding.

\section{Examples in SRD Module's Template}
\label{template}
In our Simple Relation Decoupling (SRD) module, different examples help LLMs in comprehending various descriptions of spatial relationships. To emphasize those relationships that are particularly sensitive to viewpoint changes, we select spatial relations whose interpretation can vary significantly with the observer's perspective. 
All examples we used are as follows: 
\begin{itemize} 
    \item The \{\textit{target}\} is close to the \{\textit{anchor}\}. 
    \item The \{\textit{target}\} is far away from the \{\textit{anchor}\}. 
    \item The \{\textit{target}\} is on the left of the \{\textit{anchor}\}. 
    \item The \{\textit{target}\} is on the right of the \{\textit{anchor}\}. 
    \item The \{\textit{target}\} is in front of the \{\textit{anchor}\}. 
    \item The \{\textit{target}\} is behind the \{\textit{anchor}\}. 
\end{itemize} 

\section{Sentence Matching Algorithm}
As mentioned in Section \ref{template}, we predefine an Example Set \(E\), which consists of \(k\) examples in total. Given an anchor, it returns \(k\) sentences. We employ a Sentence Matching Algorithm, as outlined in Algorithm \ref{alg:1}, to select the sentence that best aligns semantically with the original sentence.  
Specifically, after the input of the target, a single anchor, and $k$ examples, we obtained $k$ candidate sentences, denoted as \( S = \{{ S_{1}, S_{2},\ldots, S_{k}} \} \), and the original complex query is denoted as \(S_{q}\) through LLMs. All sentences are then predicted through the Target Classifier \(Clas \), to obtain pseudo-label \(\hat{P}_{q}\) and  \(\hat{P}_{i}, i \in \{1, 2, \ldots, k\}\). The matching score consists of two components. The first is the consistency score, which is the evaluation score for maintaining consistency of the pseudo-label between the generated sentence and the query. The corresponding \(S_{consist}\) is formulated as follows:
\begin{equation}\label{Sentence_Match}
\begin{aligned}
S_{consist} = \begin{cases} 1, & \text{if } \hat{P}_{q}=\hat{P}_{i}, \\ 0, & \text{otherwise}. \end{cases} 
\end{aligned}
\end{equation}

\begin{algorithm}[t]
\caption{Sentence Matching Algorithm}\label{alg:1}
\begin{algorithmic}[1] % Set the line number frequency
\STATE \textbf{Input:} Utterance $U$, Anchor set $O$, Target label $y$

% \STATE 
\STATE $N_q$ = len(U), $N_{anchor}$ = len(O)
% \STATE 
\STATE \textbf{For each} anchor \textbf{in} Anchor set $O$ \textbf{do:}
\STATE \hspace{0.5cm} \textbf{For each} example \textbf{in} Example set $E$ \textbf{do:} 
\STATE \hspace{1.0cm} query = ApplyTemplate($U$, anchor, example)
\STATE \hspace{1.0cm} sentences = LLM(query)
\STATE \hspace{1.0cm} tokens = Tokenizer($U$, sentences)
% \STATE \hspace{1.0cm} Compute similarity score by Eq.(x)
\STATE \hspace{1.0cm} Predict labels = $Clas$(token)
\STATE \hspace{1.0cm} $S_{consist}$ = GetConsistScore(Predict labels)
\STATE \hspace{1.0cm} $S_{similar}$ = GetSimilarScore(sentences,U)
\STATE \hspace{1.0cm} Score = ComputeScore($S_{consist}$,$S_{similar}$)
\STATE \hspace{1.0cm} Update the simplified sentence and Max\_Score%$S_{max}$
\STATE \hspace{1.0cm} \textbf{End for}
\STATE \hspace{0.5cm} Anchor\_LLMsentence.append(simplified sentence)
\STATE \textbf{End for}
% \STATE
\STATE \textbf{Output:} Sentence with Maximum Similarity
% \STATE \textbf{Return} Anchor\_LLMsentence
\end{algorithmic}
\end{algorithm}

To preserve the semantic integrity of the generated sentence while preventing redundant generation from LLMs, we introduce the second part of the evaluation score: the similarity score. This score primarily considers the shared word count and the lengths of the sentences. We break down both sentences into individual words and calculate the number of shared words \(N_{co\_word}\) between \(S_{i}\) and \(S_{q}\). To counteract the advantage of longer sentences in similarity calculation, we employed TextRank Algorithm\cite{mihalcea2004textrank} to compute the sentence similarity, denoted as $S_{Similar}$. Finally, the similarity score is calculated by dividing the shared word count by the sum of the logarithms of the lengths of the two sentences. The formula is as follows:
\begin{equation}\label{eq:w1}
\begin{aligned}
\text{$S_{Similar}$} = \frac{N_{co\_word}}{\log(N_q) + \log(N_s)},
\end{aligned}
\end{equation}
where \(N_{q}\) and \(N_s\) denote the length of the query and the generated sentence. In addition, we have defined an ideal decoupling sentence length \(N^*\) based on the length of the sentence and the number of anchors in the sentence. Using \(N^*\) as the center, the corresponding weights are obtained with a peak value 1, gradually decaying towards both ends. For instance, if \(N_{q}\) is 7 and there are 2 anchors, the weight probability distribution would be \{0.4, 0.6, 0.8, 1, 0.8, 0.6, 0.4\}. When \(N_{s}=2\), the corresponding weight probability \(w\) is 0.6.
Therefore, the corresponding formula for the score is as follows:
\begin{equation}\label{fig2_network}
\begin{aligned}
\text{Score }= w \cdot S_{Similar} +  S_{consist}.
\end{aligned}
\end{equation}

%------------------------------------------------------------------------
\section{Loss Function}
\textbf{Object-level Loss \(\mathcal{L}_{\text{Object}}\):}  
Followed by MVT\cite{huang2022multi}, semantic class representations \(\{L_c\}\) are obtained by encoding class name tokens with a pre-trained language encoder and extracting the [CLS] token. Given projected object features \(F' \in \mathbb{R}^{N \times D}\) (for \(N\) objects with feature dimension \(D\)) and class representations \(L \in \mathbb{R}^{C \times D}\), we compute the predicted logits as

\begin{equation}
    P_{\text{obj}} = F' \cdot L^T \in \mathbb{R}^{N \times C}.
\end{equation}

The ground truth is represented as a one-hot encoded matrix \(T_{\text{obj}} \in \mathbb{R}^{N \times C}\). The object-level loss is then defined as

% \begin{equation}
%     \mathcal{L}_{\text{Object}} = - \sum_{j=1}^{N} \sum_{c=1}^{C} T_{\text{obj}}(j,c) \, \log \!\left( \frac{\exp((P_{\text{obj}})_{jc})}{\sum_{c'=1}^{C}\exp((P_{\text{obj}})_{jc'})} \right).
% \end{equation}
% 直接使用 softmax 的写法如下：

\begin{equation}
\mathcal{L}_{\text{Object}} = - \sum_{j=1}^{N} \sum_{c=1}^{C} T_{\text{obj}}(j,c) \, \log\Bigl(\text{softmax}(P_{\text{obj}})_{jc}\Bigr).
\end{equation}

This loss ensures that the object features capture critical geometric properties, such as shape and center, and align accurately with their semantic categories.

\textbf{Referential Loss \(\mathcal{L}^{P}_{\text{ref}}\):}  
To supervise spatial alignment, followed by CoT3DRef\cite{bakr2024cot3dref}, , we obtain in parallel the anchor position labels \(\mathbf{T}_{\text{anchor}} \in \mathbb{R}^{I}\) and the target position label \(\mathbf{T}_{\text{target}} \in \mathbb{R}^{1}\), and concatenate them to form the referential ground truth

\begin{equation}
   \mathbf{T}_{\text{ref}} = \operatorname{concat}\left(\mathbf{T}_{\text{anchor}},\, \mathbf{T}_{\text{target}}\right) \in \mathbb{R}^{(I+1)}, 
\end{equation}

where each element \(T_{\text{ref},p}\) (for \(p=1,\ldots,I+1\)) indicates the correct object index (among the \(i\) objects present in the scene) for that spatial position. After view aggregation and prediction head, the model outputs referential logits \(L_{\text{ref}} \in \mathbb{R}^{(I+1) \times i}\), where each row corresponds to a spatial position (anchor or target) and each column represents one of the \(i\) objects. The referential loss is defined as

\begin{equation}
 \mathcal{L}^{P}_{\text{ref}} = -\frac{1}{I+1} \sum_{p=1}^{I+1} \log \frac{\exp\big(L_{\text{ref}, p, T_{\text{ref},p}\big)}}{\sum_{r=1}^{i} \exp\big(L_{\text{ref}, p, r}\big)}.   
\end{equation}

This loss encourages the textual module to generate discriminative features that accurately capture the semantic nuances of the grounding statement, effectively distinguishing between the anchor and target labels.

\textbf{Sentence-level Loss \(\mathcal{L}_{\text{Sent}}\):}  
Followed by CoT3DRef\cite{bakr2024cot3dref}, we supervise the textual module using a standard cross-entropy loss. Let \(I\) denote the number of anchor class labels in the original sentence. For each sample, we form a label sequence by concatenating the \(I\) anchor labels with one target label, yielding a sequence of length  
\begin{equation}
S = I + 1.
\end{equation}
After Textual Aggregation, the classification head produces language prediction logits \(\mathbf{P}_{\text{sent}} \in \mathbb{R}^{S \times C}\), where \(C\) is the total number of classes. The ground truth is represented as a vector \(\mathbf{T}_{\text{sent}} \in \mathbb{R}^{S}\), with each element \(T_{\text{sent},i}\) indicating the correct class index for the \(i\)-th position. The language loss is defined as

\begin{equation}
   \mathcal{L}_{\text{Sent}} = -\frac{1}{S} \sum_{i=1}^{S} \log \frac{\exp\big((\mathbf{P}_{\text{sent}})_{i, T_{\text{sent},i}}\big)}{\sum_{j=1}^{C}\exp\big((\mathbf{P}_{\text{sent}})_{ij}\big)}. 
\end{equation}

This loss encourages the textual module to generate discriminative features that accurately capture the semantic nuances necessary for language grounding.

% rebuttal Gs8n#Q3
\section{t-SNE visualization of CCVTS}
\begin{figure}[h!]
    \captionsetup{type=figure}
    \centering
\includegraphics[width=0.8\linewidth]{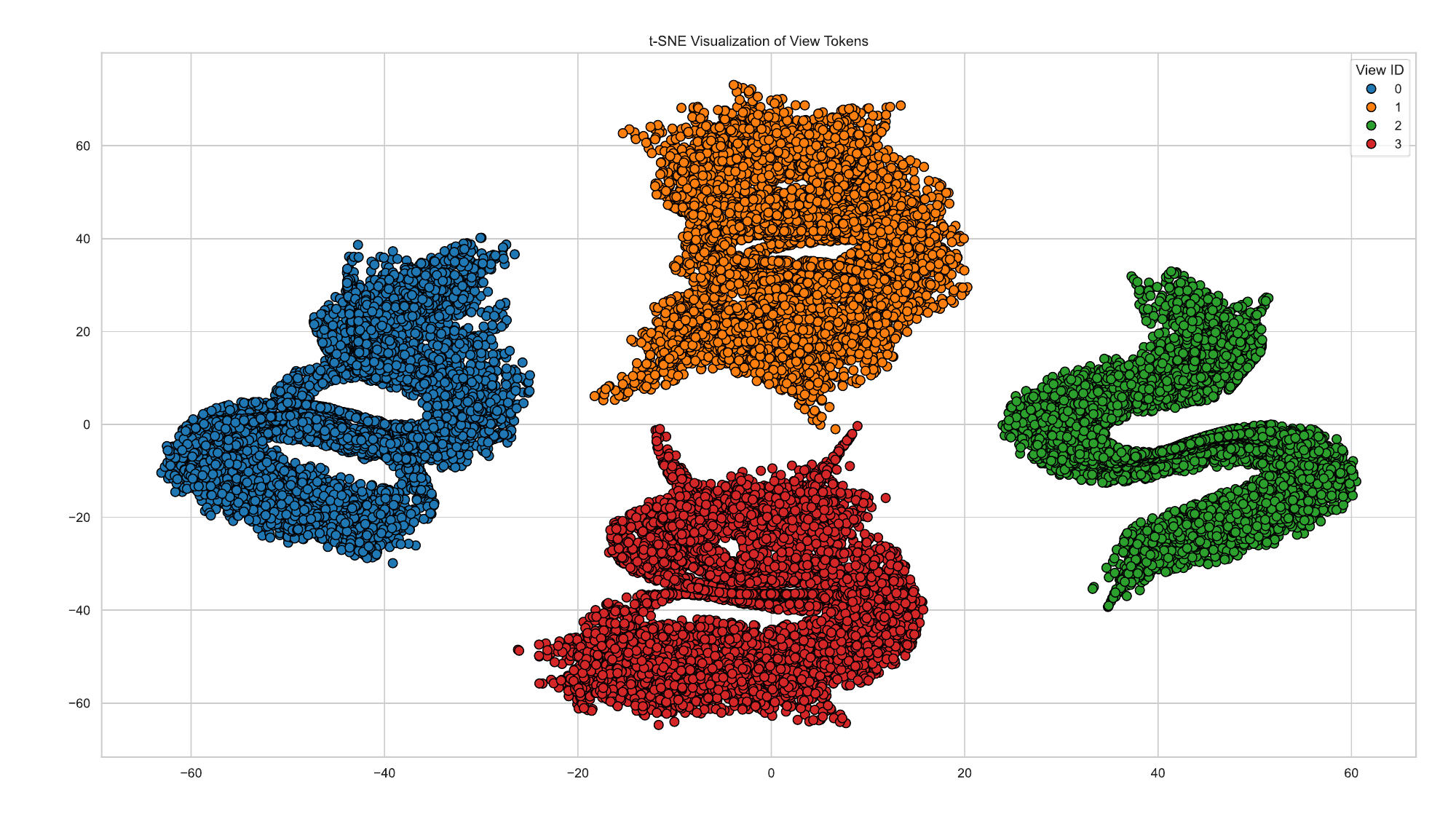}

    \vspace{-5pt}
    \caption{t-sne Visualization of CCVTs.}
    \label{fig:ccvt}
%     \hfill
%     \begin{minipage}[c]{.5\linewidth}
%             \scriptsize
%             \centering
%             \setlength{\tabcolsep}{0.1mm}
%             \begin{tabular}{lccccc}
%             \toprule
%             Method & Hard & Easy & View-Dep. & View-Indep. & All \\ \midrule
%             w/o Weight.     & 64.0 & 74.2 & 66.5 & 70.2 & 69.0 \\
%             LLM-Aug.     & 63.7 & 74.5 & 68.0 & 69.5 & 69.1 \\
%             ViewSRD   & \textbf{64.8} & \textbf{75.3} & \textbf{68.6} & \textbf{70.6} & \textbf{69.9} \\ \bottomrule
%             \end{tabular}
%         \vspace{-10pt}
%         \captionof{table}{Ablation on Nr3D.}
%         \label{tab:avg}
%     \end{minipage}
%     \vspace{-20pt}
\end{figure}
To further investigate how CCVTs encode perspective information, we visualize the learned features using t-SNE, as shown in Fig.~\ref{fig:ccvt}. Tokens originating from different viewpoints form well-separated clusters, while those from the same viewpoint are closely grouped. This clear structural separation illustrates that CCVTs effectively capture and preserve view-specific semantics, which are essential for accurate cross-view alignment in 3D visual grounding.

% rebuttal Gs8n#Q4:
\section{Analysis of Loss Weight.} 

\begin{table}[ht]
\centering
\scriptsize
\caption{Ablation on loss weights.}
\label{tab:weight}
\begin{tabular}{c c c c}
\toprule
$\lambda_{Ref}$ & $\lambda_{Obj}$ & $\lambda_{Sent}$ & Accuracy (\%) \\
\midrule
1.0 & 1.0 & 0.0 & 69.1 \\
1.0 & 0.0 & 1.0 & 63.9 \\
1.0 & 0.25 & 0.25 & 68.4 \\
1.0 & 0.5 & 0.5 & \textbf{69.9} \\
1.0 & 0.75 & 0.75 & 69.6 \\
1.0 & 1.0 & 1.0 & 69.8 \\
\bottomrule
\end{tabular}
\end{table}

Table~\ref{tab:weight} presents an ablation study on the weighting scheme of the loss components \(\mathcal{L}_{Ref}\), \(\mathcal{L}_{Obj}\), and \(\mathcal{L}_{Sent}\). The best accuracy of \textbf{69.9\%} is achieved when the loss weights are set to $\lambda_{Obj}=1.0$, $\lambda_{Ref}=0.5$, and $\lambda_{Sent}=0.5$, confirming the effectiveness of our multi-loss formulation with this specific weighting scheme.

Notably, omitting either \(\mathcal{L}_{Obj}\) or \(\mathcal{L}_{Sent}\) leads to significant performance degradation (to 63.9\% and 69.1\%, respectively), highlighting the importance of jointly modeling object-level and sentence-level supervision. We also observe that moderate weighting (e.g., 0.25 or 0.5) for \(\mathcal{L}_{Obj}\) and \(\mathcal{L}_{Sent}\) improves over their absence, but still underperforms compared to the balanced setting. These results validate that all three components contribute synergistically to grounding performance.

\end{document}